\documentclass[runningheads]{llncs}
\usepackage[T1]{fontenc}
\usepackage{graphicx}
\usepackage{multirow}
\usepackage{hyperref}
\usepackage{array}
\usepackage{cite}
\usepackage{enumitem}
\usepackage{caption}
\usepackage{subcaption}
\usepackage{xcolor}
\begin{document}
\title{Assorted, Archetypal and Annotated Two Million (3A2M) Cooking Recipes Dataset based on Active Learning }
\titlerunning{3A2M Cooking Recipes Dataset based on Active Learning}
% If the paper title is too long for the running head, you can set
% an abbreviated paper title here
%
\author{Nazmus Sakib\inst{1} \and
G. M. Shahariar\inst{2} \and
Md. Mohsinul Kabir\inst{1} \and
Md. Kamrul Hasan\inst{1} \and
Hasan Mahmud\inst{1}}
\authorrunning{Sakib et al.}
% First names are abbreviated in the running head.
% If there are more than two authors, 'et al.' is used.
%
\institute{Islamic University of Technology (IUT), Gazipur, Bangladesh \and
Bangladesh University of Engineering and Technology, Dhaka, Bangladesh
\email{\{nazmussakib009,sshibli745\}@gmail.com, \{mohsinulkabir, hasank, hasan\}@iut-dhaka.edu}}
\maketitle              % typeset the header of the contribution
\begin{abstract}
Cooking recipes allow individuals to exchange culinary ideas and provide food preparation instructions. Due to a lack of adequate labeled data, categorizing raw recipes found online to the appropriate food genres is a challenging task in this domain. Utilizing the knowledge of domain experts to categorize recipes could be a solution. In this study, we present a novel dataset of two million culinary recipes labeled in respective categories leveraging the knowledge of food experts and an active learning technique. To construct the dataset, we collect the recipes from the RecipeNLG dataset \cite{bien2020cooking}. Then, we employ three human experts whose trustworthiness score is higher than 86.667\% to categorize 300K recipe by their Named Entity Recognition (NER) and assign it to one of the nine categories: bakery, drinks, non-veg, vegetables, fast food, cereals, meals, sides and fusion. Finally, we categorize the remaining 1900K recipes using Active Learning method with a blend of Query-by-Committee and Human In The Loop (HITL) approaches. There are more than two million recipes in our dataset, each of which is categorized and has a confidence score linked with it. For the 9 genres, the Fleiss Kappa score of this massive dataset is roughly 0.56026. We believe that the research community can use this dataset to perform various machine learning tasks such as recipe genre classification, recipe generation of a specific genre, new recipe creation, etc. The dataset can also be used to train and evaluate the performance of various NLP tasks such as named entity recognition, part-of-speech tagging, semantic role labeling, and so on. The dataset will be available upon publication: \url{https://tinyurl.com/3zu4778y}. 

\end{abstract}
\vspace{-5mm}
\section{Introduction}

A recipe is a text that is used when cooking or baking food that specifies the cooking method, the items required, and how to use them. Recipes allow individuals to make new dishes without having to watch a demonstration. One maybe familiar with the basics of food preparation, but the preparation of certain meals, such as Sushi and Baklava, have unique procedures that must be learned before they can be properly cooked. Even if a person has a large number of ingredients in their refrigerator, making some recipes can be challenging. A recipe can save a lot of his time by instructing him on how to prepare the food, which ingredients to use, how to prepare them, and any nutritional information that may be relevant. A person may even find a recipe on the internet for a dish that he or she has never heard of before. As recipes describe a set of rules for cooking food in an informal way, there is no strict rule of how these texts should be structured. The same recipe may appear in many journals and multiple periodicals in different representations, but are actually the same in outcome. Recipes can also be organized by nation or continent \cite{bien2020cooking, fisher1969anatomy}. In this skewed circumstance, if the meal is evaluated in a certain genre, it can allow consumers to make an informed choice based on their interests and tastes \cite{recipes2022food}. It is also worth noting that, while each meal (depending on its ingredients) may fall into a specific genre, each dish has the capacity to stand alone or fall into another. A salad, for example, can be a salad, a side dish, a main meal, or a dessert. The same holds true for a main course. 

There is a surge of interest in using culinary recipe datasets for deep learning research \cite{lecun2015nature}. This is because culinary recipes include a wealth of data that can be utilized to train deep learning models. There are, however, a scarcity of publicly available culinary recipe datasets appropriate for training deep learning models. The RecipeNLG dataset \cite{bien2020cooking} alleviate this issue by including both culinary recipes and named entities of more than 2 million food items. The corpus includes culinary recipes from various sources such as cookbooks, blogs, and recipe websites. It is regarded as the first publicly accessible dataset of culinary recipes. This dataset was inspired by the Recipe1M+ dataset \cite{marin2019dataset}. RecipeNLG piqued people's attention in the issue by having the most publicly available recipe collection at the time. Though there are no hard and fast rules for classifying global cuisines into genres, we can say that there is a distinct split of food market customers based on the components utilized or the nature of the meal patterns. The most prevalent distinction is between "fast food" and "slow food" marketplaces [38]. According to our findings, the RecipeNLG \cite{bien2020cooking} dataset's recipes are not classified into any genres. In this study, we split two million recipes into nine categories, each with its own set of justifications based on domain experts' judgments \cite{price2020six}.

This study aims to contribute in this domain through: (1) constructing a recipe dataset that contains nine genres developed through domain experts, (2) applying active learning and ensemble-based techniques to semi-automate the annotation process of 2 million data using Human-in-the-loop  approach. The outcome of this study is an annotated original dataset of two million culinary recipes (3A2M Cooking Recipes Dataset), as well as the possibility of new recipes being developed utilizing this information, allowing people to select food meals based on their favored categories. We began by collecting recipes from the RecipeNLG dataset. Three human experts were employed, whose trustworthiness score is higher than 86.667\% to categorize 300K recipe by their Named Entity Recognition (NER) and assign it to one of the nine categories: bakery, drinks, non-veg, vegetables, fast food, cereals, meals, sides and fusion. The remaining 1900K recipes were then categorized using Active Learning and a query by committee procedure. The purpose of this study is to emphasize the diversity of the culinary genre in multi-class so that people may learn more about it and develop their abilities. It also tries to emphasize that the culinary genre comprises activities other than cooking, such as baking, pastry making, confectionery manufacturing, and beverage preparation. We believe that the research community can utilize this dataset to train more robust language models and can perform culinary related machine learning tasks such as recipe genre classification, recipe generation of a specific genre, new recipe creation, etc.

The rest of the paper is organized as follows. Some previous related works are presented in section ~\ref{sec:RW}. Section ~\ref{sec:RD} contains the description of the RecipeNLG dataset. In section ~\ref{sec:DC}, we have discussed the 3A2M corpus annotation procedure and dataset validation procedure in detail. Section ~\ref{sec:concusion} concludes this study with some future research directions.

\section{Related Work}
\label{sec:RW}
Recipe datasets have been in the periphery of the research community's attention due to their richness in linguistic elements. Several research articles on recipe collection and preparation have been published in recent years. Though traditional tactics were beneficial in some studies, there is a lot of scope in this domain when it comes to resources and frameworks. Recipe1M+ dataset\cite{marin2019dataset} is a pioneering work that inspired many culinary studies later on. Recipe1M+ is a large-scale, organized corpus containing over one million culinary recipes and 13 million food photos. It provided the opportunity to train high-capacity models on aligned, multi-modal data because it was the biggest publicly available collection of recipe data until 2020. Using this data, they trained a neural network to learn a combined embedding of recipes and photos, which produced outstanding results on an image-recipe retrieval test. They proved in that study that regularization with the addition of a high-level classification target both increases retrieval performance to rival that of humans and allows semantic vector arithmetic. A lot of works used the Recipe1M+ dataset to develop further resources. For example, using the Recipe1M+ dataset \cite{marin2019dataset} and a language model, Lee et al. \cite{lee2020recipegpt:} developed a method for autonomously producing culinary recipes. Translation metrics were used to evaluate the model. They concentrated on two tasks: component development and instruction development. Their findings demonstrated that the language model was capable of producing recipes that were semantically comparable to the original recipes. Culinary research through data-driven methods has been hindered by a scarcity of publicly available culinary recipe datasets for machine learning models \cite{marin2019dataset}. The RecipeNLG dataset \cite{bien2020cooking}, which includes both culinary recipes and named entities, might be viewed as a prominent contribution for attempting reduction of this scarcity. The corpus includes culinary recipes from various sources such as cookbooks, blogs, and recipe websites. The RecipeNLG authors have witnessed a significant increase in interest in utilizing culinary recipe datasets for deep learning studies. They were motivated by the release of the Recipe1M+ dataset, though it included both recipes and photos. They employed a Named Entity Recognizer (NER) to extract food entities from the dataset and feed them into the recipe generator through specific control tokens. This information is used to fine-tune a GPT-2 language model, which creates new recipes based on a given set of food things. On top of the Recipe1M+ dataset, the new dataset includes over 1 million fresh, preprocessed, and deduplicated recipes. They broaden the scope to allow normalization to a given number of servings. Another intriguing possible task may be the unification of primarily ambiguous units in relation to the item they are describing, which could have numerous applications in and outside of the culinary realm, and further unification using knowledge graphs. Though these recipe datasets made significant progress in culinary research through data-driven methods, there is a lack of large scale annotated recipe dataset based on genre which can help training robust machine learning models to foster research in this domain. When it comes to annotate unlabeled data, active learning is a commonly used method in multidisciplinary domain. Active learning is a supervised machine learning technique that trains a predictor iteratively and utilizes the predictor to pick the training instances in each iteration, boosting the predictor's odds of selecting better configurations and improving the prediction model's accuracy \cite{settles2009active}. The Naive Bayes (NB) \cite{hastie2009elements} classifier is commonly used as an active learning predictor. It is predicated on Bayes' theorem and strong (naive) independent assumptions about features. To categorize unlabeled data, supervised machine learning methods such as Support Vector Machine (SVM) \cite{cortes1995support-vector}, Logistic Regression \cite{hastie2009elements}, and Random Forest (RF) \cite{ho1995random} are also employed as predictors. However, they are usually used to overcome classification problems. At each level, multi-layer perceptrons (MLP)\cite{hastie2009elements} learn to abstract and mix data. The data is processed in several layers. Back propagation is utilized to train such neural network models with several layers \cite{mitchell1997machine}. Active learning is utilized in many different applications, such as image classification, speech recognition, and text categorization. Rather than assuming that all of the training examples are supplied at the start, active learning algorithms gather more instances on the fly, often by asking a human user questions. The query by committee (QBC) method is an example of an active learning algorithm since it picks a selection of training samples that are most likely to enhance the classifier's performance. Unlabeled data searches are usually a mix of semi-supervised and active learning scenarios. This method is used to classify unlabeled data and keep a high-quality training dataset\cite{cohn1996active}. In this study, we leverage the data from RecipeNLG dataset along with active learning and query by committee method to create a large scale annotated culinary dataset. 

\section{RecipeNLG Dataset}
\label{sec:RD}
%\vspace{-0.98em}
The RecipeNLG collection is the largest accessible dataset in the domain, with 2,231,142 different culinary recipes. One of the shortcomings  of this dataset is that the genre of the recipes were discovered to be not categorized or unclassified. A culinary recipe's structure includes a title, a list of ingredients with quantities, and step-by-step directions. The title, the simplest component of the recipe, correctly identifies and explains the contents. Some examples from the RecipeNLG dataset are depicted in table ~\ref{tab:tab00}.

%\vspace{-3em}
\begin{table}[!ht]
\setlength{\belowcaptionskip}{-15pt}
\caption{Example samples from RecipeNLG dataset \cite{bien2020cooking}}
\label{tab:tab00}
\centering
\begin{tabular}{|c|c|c|}
\hline
\textbf{Title}                                                       & \textbf{Directions}                                                                                                                                                                                                                                                    & \textbf{NER}                                                                                                                                           \\ \hline
\begin{tabular}[c]{@{}c@{}}Chocolate \\ Frango \\ Mints\end{tabular} & \begin{tabular}[c]{@{}c@{}}{[}"Mix ingredients together for 5 minutes.", \\ "Scrape bowl often. \\Last fold   in chocolate chip mints.", \\ "Bake at 350\textbackslash{}u00b0 for 35 to 40 minutes   \\ or until done (cake mix directions)."{]}\end{tabular}            & \begin{tabular}[c]{@{}c@{}}{[}"cake mix", \\ "chocolate fudge pudding",\\  "sour cream", "water", \\ "Wesson oil", \\ "eggs", "Frango"{]}\end{tabular} \\ \hline
\begin{tabular}[c]{@{}c@{}}Cold \\ Spaghetti \\ Salad\end{tabular}   & \begin{tabular}[c]{@{}c@{}}{[}"Cook noodles 1/2 done (do not rinse).", \\ "Add all other ingredients; mix   well.", \\ "Marinate at least 3 hours.", \\ "The longer it   sits, the better it gets!!"{]}\end{tabular}                                                   & \begin{tabular}[c]{@{}c@{}}{[}"vermicelli", "bell pepper", \\ "zucchini",\\  "purple onion", \\ "Salad Supreme seasoning"{]}\end{tabular}              \\ \hline
\begin{tabular}[c]{@{}c@{}}Chocolate \\ Pie\end{tabular}             & \begin{tabular}[c]{@{}c@{}}{[}"Mix dry ingredients; add milk.", \\ "Beat in egg yolks.", \\ "Pour on 1 cup boiling water.", \\ "Mix and cook until thick.",   \\ "Cool.", \\ "Pour in baked pie shell.", \\ "Cover with   meringue and brown in oven."{]}\end{tabular} & \begin{tabular}[c]{@{}c@{}}{[}"sugar", "milk", "cocoa", \\ "flour", "egg yolks",\\  "clump", "vanilla", \\ "boiling water"{]}\end{tabular}             \\ \hline
\end{tabular}
\end{table}
\noindent All ingredients are proportionate to the serving size of the dish. The quantity is linked to the name of the unit. The processes involved in preparing certain food products are listed in the directions feature. Every ingredient on the list is utilized in the correct amounts. However, it was reported that the RecipeNLG dataset has a few limitations in terms of the validity of the recipe title structure \cite{yagcioglu2018recipeqa:}. The authors of RecipeNLG drew inspiration from the 1M+ Recipes dataset and contrasted it. They deleted duplicate recipes and translated them all into English. Though they did not discover the unified matrices, they did discover a list of substances known as Named Entity Recognition (NER), which is not exhaustive because the same item appeared in many recipes. It enables the employment of ensemble learning approaches by categorizing the data with suitable categories \cite{bien2020cooking} \cite{marin2019dataset} \cite{pods97}.

\section{3A2M Dataset Construction}
\label{sec:DC}
This section describes in detail the 3A2M corpus development, annotation, and assessment techniques. The overall dataset construction process is shown in Figure ~\ref{process}. 
\subsection{Corpus Description}
The 3A2M dataset is built on top of the RecipeNLG dataset and we have incorporated as well as utilized all the data along with their respective features. In terms of recipe name, cooking technique, ingredients, and recipe sources, all of the data are the same. 3A2M dataset has in total five attributes: \textit{title, directions, NER, genre}, and \textit{label} among which the data of \textit{title, directions}, and \textit{NER} attributes are directly incorporated from RecipeNLG dataset. Random 300K recipes are classified to one of nine categories by three human experts whose trustworthiness score is more than 86.667 percent based on the related Named Entity Recognition (NER): \textit{bakery, drinks, non-veg, vegetables, fast food, cereals, meals, sides}, and \textit{fusion}. The remaining 1900K recipes are automatically classified using active learning and a query by committee approach. As a result, there are more than two million recipes, each of which is classified into a certain genre and assigned a confidence score. The huge dataset's Fleiss Kappa score for the nine genres is around 0.56026. The techniques for evaluating corpora are covered in a subsequent section. To preprocess the original unlabeled data, Natural Language Processing (NLP) techniques such as unique word discovery, genre principle categorized word matching, and lowercase English letter conversion are employed.

\subsection{Corpus Annotation Procedure}
\noindent Predictors are data-driven algorithms that utilize data to forecast occurrences or infer a function. In a classification problem, we instead attempt to predict outcomes given a collection of distinct inputs. To put it another way, we are attempting to categorize input variables. Instead of delivering the intended result, we may transform the recipe data into a categorization challenge. In the 3A2M dataset, the recipes are organized into nine categories. Experts have identified nine separate genres, each with its own set of food entities or ingredient constraints. Food professionals determine the food specifications. Though we are familiar with the words "continental cuisine," "eastern cuisine," and "north american genre", several food sectors, however, create food zones based on their own preferences \cite{kiddon2016}\cite{khodak2018large}.
\begin{figure}
\includegraphics[width=\textwidth]{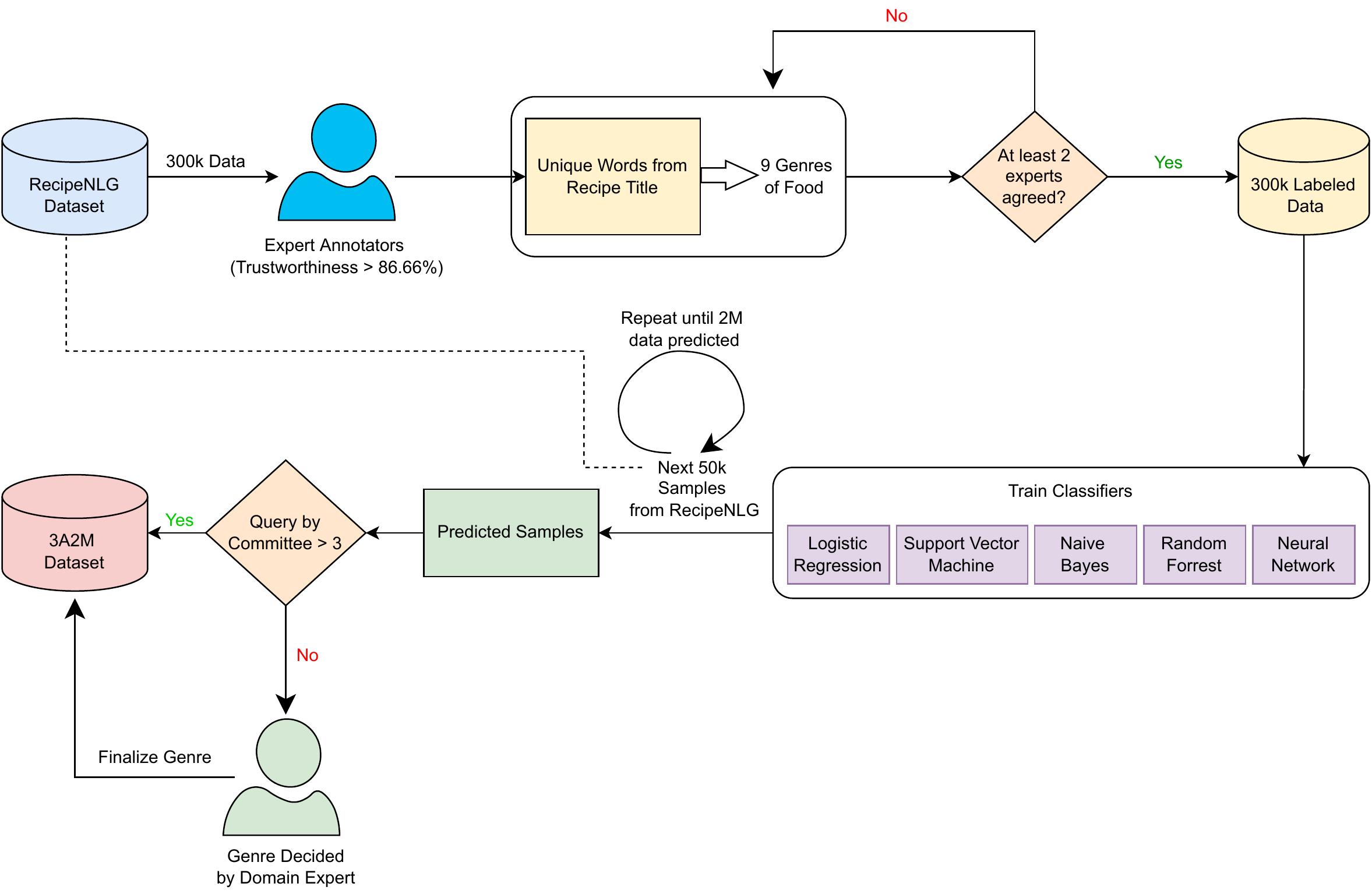}
\setlength{\belowcaptionskip}{-15pt}
\caption{Working Procedure of 3A2M Cooking Database} \label{process}
\end{figure}
The working procedure of the dataset annotation is illustrated in Figure ~\ref{process}. We have divided the corpus annotation into two steps. The first step is genre identification and initial data labeling by expert human annotators. The second step is to use active learning to automatically label the remaining unlabeled data using query by committee strategy and human in the loop approach. Each step is described in detail. First and foremost, we needed to determine the number of genres. To do this, we went through a large number of books and periodicals and determined that cuisines are typically classified by zone, such as Asian, South Asian, European, Arabian, Indian, Chinese, Japanese, Lebanese, or Italian \cite{wiki} \cite{eathappyproject.com}. With such a diverse range of cuisines in this type of genre, it might be difficult to establish which category a particular item belongs in. Food may become famous in one place, but its origins might be traced back to another. The French, for example, may be credited with inventing the hamburger, although the hamburger was created in the United States \cite{yang2017reference-aware} \cite{salvador2017learning}.
\begin{table}[!ht]
\setlength{\belowcaptionskip}{-5pt}
\caption{ Genre id, Genre Name and Facts and Instances in 3A2M Dataset.}
\label{tab:tab01}
\centering
\begin{tabular}{|c|c|c|c|}
\hline
\textbf{\begin{tabular}[c]{@{}c@{}}Genre\\ ID\end{tabular}} & \textbf{\begin{tabular}[c]{@{}c@{}}Genre\\ Name\end{tabular}} & \textbf{Description}& 
\textbf{\begin{tabular}[c]{@{}c@{}}No. of \\ instances\end{tabular}} \\ \hline
1                                                           & Bakery                                                        & \begin{tabular}[c]{@{}c@{}}This area mostly contains baked or fried \\ foods that are served in the open or may \\ be stored for a long period.\end{tabular}                                                                                      & 160712                                                               \\ \hline
2                                                           & Drinks                                                        & \begin{tabular}[c]{@{}c@{}}Drinks are in the liquid zone and can be \\ blended with any chemical or ionic drink\\ in this zone.\end{tabular}                                                                                                      & 353938                                                               \\ \hline
3                                                           & NonVeg                                                        & \begin{tabular}[c]{@{}c@{}}This zone includes foods such as curries \\ of poultry, beef, and fish, which can be \\ self-serving or mixed serving.\end{tabular}                                                                                    & 315828                                                               \\ \hline
4                                                           & Vegetables                                                    & \begin{tabular}[c]{@{}c@{}}Foods cooked differently than the meats, \\ seafood, and eggs found in this zone. \\ This zone was built just for vegetarians.\end{tabular}                                                                            & 398677                                                               \\ \hline
5                                                           & Fast Food                                                     & \begin{tabular}[c]{@{}c@{}}Only quick food is baked or fried food that\\  cannot be kept for an extended period of\\ time in an open or cold environment. \\ Fast food may be produced by combining \\ bakery and cereal goods.\end{tabular}      & 177109                                                               \\ \hline
6                                                           & Cereal                                                        & \begin{tabular}[c]{@{}c@{}}Cereals are mainly foods made from corn, \\ wheat, and rice. We have placed meals that \\ are directly generated from grains in the \\ corn zone.\end{tabular}                                                         & 340495                                                               \\ \hline
7                                                           & Meal                                                          & \begin{tabular}[c]{@{}c@{}}Some items may appear to be quick food, \\ yet they might actually constitute a complete \\ meal. Nonveg/vegetable products, like \\ platters, can also be served with cereals \\ to provide a full meal.\end{tabular} & 53257                                                                \\ \hline
8                                                           & Sides                                                         & \begin{tabular}[c]{@{}c@{}}The medicines, sauces, and toppings are \\ basically sections in the side section.\end{tabular}                                                                                                                        & 338497                                                               \\ \hline
9                                                           & Fusion                                                        & \begin{tabular}[c]{@{}c@{}}Some food that can be properly sorted. \\ Sometimes experts disagree on whether \\ it belongs in a specific category known as \\ fusion meals.\end{tabular}                                                            & 92630                                                                \\ \hline
\end{tabular}
\end{table}
Taking into account these stressful scenarios, we took help from three expert annotators who have experience working with food recipes. We evaluated their trustworthiness score and all three of them have a trustworthiness score higher than 86.667\%. Those three experts categorized meals in a novel way in the 3A2M dataset by forming nine unique genres, each with its own set of culinary goods or components norms. The genres are listed in table ~\ref{tab:tab01}. Initially, the experts intended to collaborate by using the NER associated with each recipe in the RecipeNLG dataset, but their significant expertise revealed that many ingredients had duplication in the same dish and that the same components were utilized in various genres. To address this, 100K data points were picked at random from a pool of 2231142. They utilized the food item's title and listed the unique phrases. They detected around 4329 unique terms. However, many of the terms in the unique word list are not related to food; others are proper nouns of restaurants, persons, chefs, or locales. So they had to examine a lexicon to determine whether these terms were related to food \cite{cuisine1} \cite{dictionary25}.
They specifically found 914 words that are directly connected to eating. To fit in with the provided words, a list of keywords with the appropriate genre is constructed. Following that, each expert voted on each of the genre classification recipes. Each dish earned one of three levels of confidence: 33.33 percent, 66.667 percent, and 100.00 percent. A perfect score was given to 68,236 of the 100K food titles. The remaining 28,729 titles received 66.66\%, while the remaining 3035 titles received 33.33\%. As a consequence, 3035 titles had to be re-evaluated and finalized by examining the cooking directions. Finally, 391 titles formed a new genre known as "fusion." Finally, the procedure was repeated, and a total of 300,000 sample points were initially categorized into 9 genres. The second step is presented in the following subsection.

\subsection{Use of Active Learning}
Machine learning methods are used to annotate 2 million recipes in order to grasp the current status of implementation. This study used traditional machine learning classifiers to classify items as bakery, beverages, non-veg, vegetables, fast food, cereals, meals, sides, or fusion. Three skilled annotators classified 300K dishes into genres in the first step. The remaining unlabeled data is labeled using active learning. To increase the efficiency of the data labeling process, changes are made to the active learning process. To categorize the remaining 1 million and 900K data points, we utilized five machine learning classifiers: Logistic Regression, Support Vector Machine (SVM), Naive Bayes (NB), Multi-layer Perceptron (MLP) and Random Forest (RF). It is conceivable for a learning method to perform well in one matrix while being sub-optimal in another. To avoid classifier bias, considering the accuracy, sensitivity, specificity, and precision parameters of classifiers might be a solution. An ensemble of classifiers can also be utilized to improve the performance of a single classifier. In the active learning process, the ensemble of classifiers may be utilized to decide on the class of a certain recipe, which is used as an input for the active learning process. In this scenario, we utilized the majority vote of all classifiers \cite{kalchbrenner2014convolutional}. We first trained these five classifiers with 300K labeled recipe titles. After that, in each active learning iteration, we randomly chose 50K titles from the dataset and let the classifiers predict corresponding genres. The technique follows the Query by Committee procedure \cite{settles2009active}, so if the result indicated more than three classification algorithms categorizing a work in a given genre, that label was accepted and those 50K instances were added to the initial labeled dataset. That indicates that the confidence score is always greater than 60\%. Otherwise, for the instances in which we could not fix the genre by following the query by committee procedure, we used the human in the loop (HITL) procedure to manually fix the genre of those specific instances and add them into the labeled training dataset. The whole procedure was repeated until all the unlabeled instances were labeled. After each iteration, the predictor models were trained using the updated labeled dataset.\\

\begin{figure}
\includegraphics[width=\textwidth]{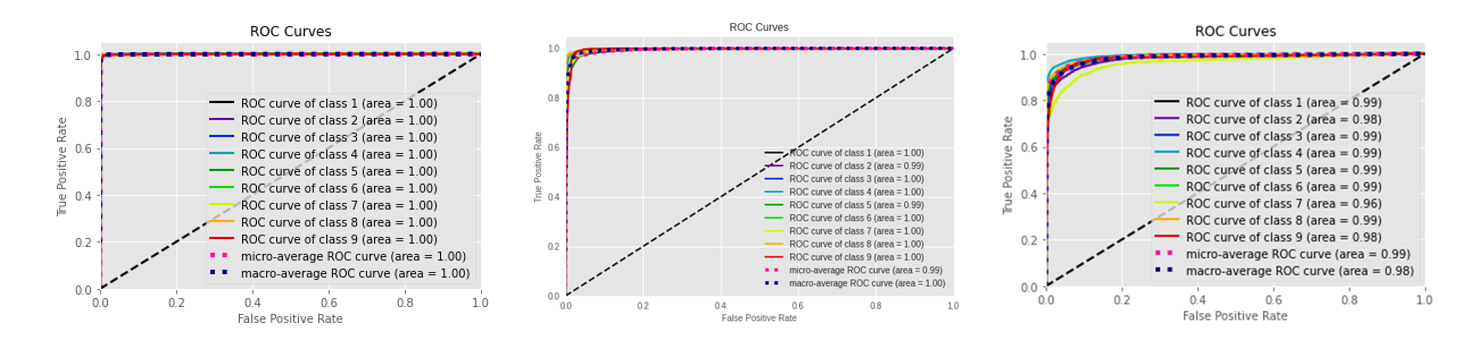}
\setlength{\belowcaptionskip}{-15pt}
\caption{ROC Curve for Logistic Regression,  NB and RF} \label{fig:roc}
\end{figure}

\begin{figure}
\includegraphics[width=\textwidth]{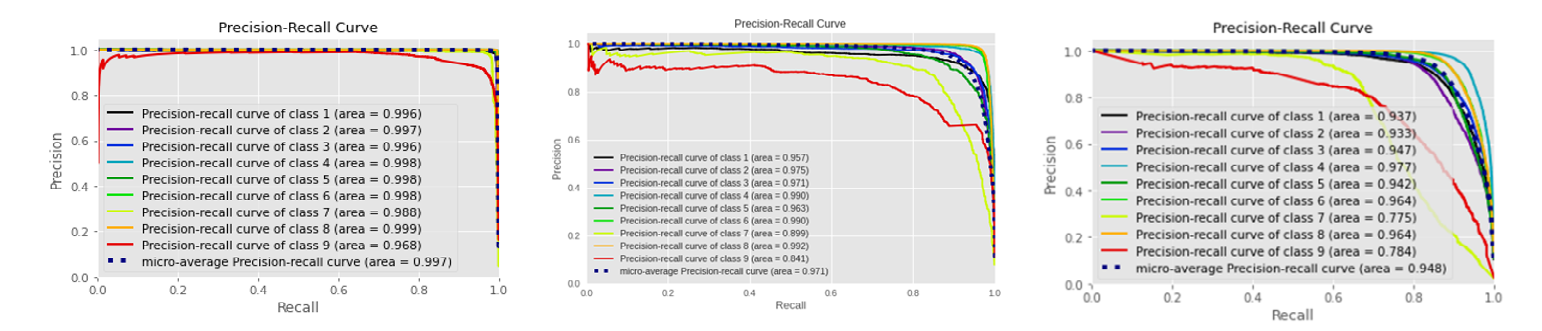}
\setlength{\belowcaptionskip}{-15pt}
\caption{Precision Recall Curve for Logistic Regression, Naive Bayes, Random Forest.} \label{fig:recall}
\end{figure}

\noindent The optimum cutoff value for predicting new genres is determined using ROC (Receiver Operating Characteristic) curves. Instead of explicitly predicting classes, the precision recall curve shows versatility in evaluating the likelihood of an observation belonging to each class in a classification problem. Figures ~\ref{fig:roc} and ~\ref{fig:recall} show the ROC curve and the Precision Recall Curve of Logistic Regression, Naive Bayes, Random Forest classifiers, respectively.

\begin{figure}
\centering
\includegraphics[width=.8\textwidth]{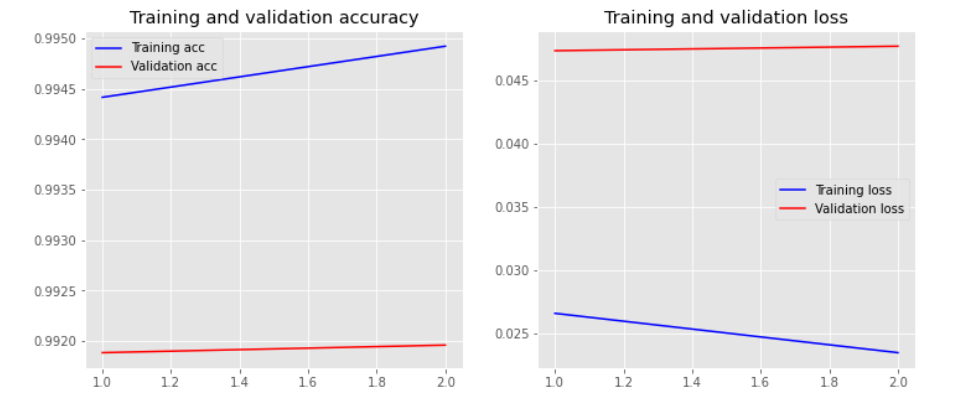}
\setlength{\belowcaptionskip}{-15pt}
\caption{ Multi-layer Perceptron training and validation accuracy and loss graph}
\label{fig:validation}
\end{figure}

\noindent A learning curve derived from the training dataset that indicates how effectively the model is learning. It is typical to generate dual learning curves for a machine learning model during training on both the training and validation datasets. The model's loss is nearly always lower on the training dataset than on the validation dataset. Figure ~\ref{fig:validation} depicts how the fit is detected by a training and validation loss that reduces to a point of stability with a limited gap between the two final loss values.

\subsection{Corpus Evaluation}
This section presents the statistical reliability measures to evaluate the quality of the annotated corpus . 
\subsubsection{Inter-Rater Reliability (IRR)} \cite{stats}: The amount of agreement amongst experts is referred to as inter-rater reliability (IRR). We calculated the IRR value from 300K of expert-annotated data. If three experts agree, the IRR is 100\%; if everyone disagrees, the IRR is 0\%. We have simulated the agreement and found an IRR value of 50.3976667\% for the 300K data over 9 genres. As there are multiple class levels, the IRR value is lower \cite{phalippou2008hazards}.
\subsubsection{Fleiss Kappa} \cite{fleiss-kappa-excel}: Fleiss Kappa is a method of determining the degree of agreement among three or more raters when they are assigned category ratings to a group of objects. For example, if the raters were asked to rate the genre of a series of recipes, expert agreement would be evaluated as the proportion of recipes rated the same by each. Although there is no formal technique for interpreting Fleiss' Kappa, the values 0.20 = "poor," 0.20 = "fair," 0.41 = "moderate," 0.61 = "good," and 0.81 = "very good" show how the number is used to quantify the level of inter-rater agreement among raters. After computing over 300K annotated data by professionals, we identified the Fleiss' Kappa value of 0.4965 in the moderate zone. After processing the data over 300K, we identified a Kappa value of 0.5 in the strong zone since there are 9 genres. When the class numbers are large, the Kappa value of the strong zone is greater than that of the moderate zone \cite{falotico2015fleiss}.
\subsubsection{Trustworthiness Score} \cite{elo2014}: The quality of a categorized dataset is determined by the annotators. There are numerous approaches to calculate the trustworthiness score of an annotator. For example, the annotators may be asked to rate how well they found each sentence on a scale of 0 to 100. In this study, for calculating annotators' trustworthiness score, we randomly selected 470 labeled data samples from the dataset and created 30 control samples. Control samples are chosen by domain experts. The control samples were easy to understand and classify. For example, tea is a popular beverage across the world. The annotators were not aware of these control samples. We found the trustworthiness scores of the three annotators were 86.667\%, 90\%, and 96.667\%, respectively, based on their responses to the control samples. As the set of classes was 9, domain experts felt that more than 80\% would be adequate to qualify the annotators. In our case, the trustworthiness score of each annotator is above 80\%, and the average trustworthiness score is 90\%, which indicates that the annotators are qualified to annotate.
\subsubsection{Confidence Score} \cite{mandelbaum2017distance-based}: A confidence score is essentially a measure of annotative quality. A confidence score can be used to exclude low-quality annotations, but it is not a suitable overall measure of annotation quality. It depicts the amount of agreement across numerous annotators; that is, weighted by the annotators' trustworthiness ratings; and it demonstrates our "confidence" in the result's validity. The aggregate result is determined based on the most confident response. In the event of a tie, we select the result with the most annotators.  We calculated confidence scores of 100\% for 89,378 recipes and 66.667\% for 201,982 recipes across 300K entries. After conferring with the three annotators, the domain expert, a chef who has been cooking for more than 16 years, solved the ties. We then employed the domain experts to solve the remaining situations. The domain experts answered all of the remaining instances, and the overall confidence score for all of the recipes was 100\%.
\begin{figure}[!ht]
\centering
\includegraphics[width=.8\textwidth]{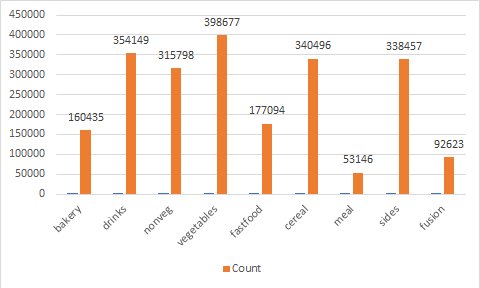}
\setlength{\belowcaptionskip}{-15pt}
\caption{3A2M Dataset Distribution}
\label{fig:distribution}
\end{figure}
\subsubsection{Anderson-Darling Test (AD Test)} \cite{razali2011power}: The Anderson-Darling test, which is dependent on the statistic, was employed for our dataset. It's similar to the t-value in t-tests or the F-value in F-tests. We do not usually interpret this statistic directly, but it is used by the program to generate the p-value for the test. We have an AD test Result on 100,000 data distribution. Figure !\ref{fig:distribution} shows the distribution of the data on 9 genres by histogram reflection.
%\vspace{-2em}
\begin{table}[]
\caption{Anderson-Darling Statistical analysis of the dataset.}
\label{tab:tab02}
\centering
\begin{tabular}{|c|c|}
\hline
\textbf{Criterias} & \textbf{Values}   \\ \hline
Sample Mean        & 4.162             \\ \hline
Sample Sigma       & 2.381             \\ \hline
AD test statistic  & 2.5937            \\ \hline
AD* test statistic & 2.613899          \\ \hline
P-value            & \textless{}0.0005 \\ \hline
\end{tabular}
\end{table}
Table ~\ref{tab:tab02} shows the AD test statistics of the Lower p-values indicate greater evidence against the null hypothesis. A p-value is the chance that the effect seen in our sample would occur if the null hypothesis for the populations were true. P-values are derived using our sample data and the null hypothesis. Lower p-values suggest more evidence against the null hypothesis, in contrast to our p-value with the degree of significance. If the p-value is less than the significance level, we reject the null hypothesis and conclude that the effect is statistically significant. In other words, the evidence in our sample is strong enough to reject the null hypothesis at the population level.

\subsection{Discussion}
To automate the process of labeling the unlabeled data, we have used active learning with query-by-committee approach. For the committee, we have used five weak machine learning classifiers as the learners. Support Vector Machine (SVM) had "linear" kernel as the parameter and Multinomial Naive Bayes had the additive laplace smoothing parameter value of 1. Random Forest classifier was initialized with 100 estimators, gini impurity, and a max-depth of 50. The penalty and class weight parameters in Logistic Regression were set to "l2" and "balanced respectively. On the other hand, Mulit-layer Perceptron was initialized with 200 max iteration, hidden layer size was 100, learning rate parameter was set to "constant" and "adam" optimizer was used.  The pre-trained model learns broad characteristics like language and context. As a result, the pre-learning model's embedding results lack adequate features differentiating the labels of downstream tasks. Therefore, data with the same label might have different features in a pre-trained model \cite{kim2022effective}. We could have incorporated large pre-trained language models like BERT by fine-tuning with the help of Transfer Learning, but due to resource constraints, the cost of training some layers of a such large model as a learner inside the active learning procedure, and for the sake of the simplicity, we did not consider BERT for labeling purpose in our study. The purpose of corpus assessment study is to obtain an overall measurement of the performance of a statistical technique and its data, which is the 3A2M dataset. Experts' assessments of circumstances and occurrences naturally differ. The IRR is mild, with the goal of minimizing subjectivity as much as feasible in our study. To ensure that the annotators were fit for this job, we measured trustworthiness score and calculated Fleiss Kappa score to measure the agreement among them. Our research demonstrates that corpus evaluation is moderate in the sense that the size of the data was greater than 2 million, and dependability demonstrates consistency of a measure by different methodologies such as confidence score, anderson-darling test.

\section{Conclusion and Future Works}
\label{sec:concusion}
Construction of food recipe dataset is a kind of tedious job which requires careful inspection. Specially for nearly matched ingredients for two or more categories. Taking help from domain experts and integrate their knowledge though machine learning approach is the worth things to do to generate a reliable dataset. Construction of food recipe dataset is a kind of tedious job which requires careful inspection. Specially for nearly matched ingredients for two or more categories. Taking help from domain experts and integrate their knowledge though machine learning approach is the worth things to do to generate a reliable dataset. Therefore, we have constructed the (3A2M) cooking recipes dataset, an annotated dataset of two million culinary recipes so that people can choose the food recipes according to their preferred categories. We built this dataset on top of the RecipeNLG dataset. First, we categorized 300K recipes into nine categories by human annotators and trained five machine learning classifiers to employ active learning for automatically labeling the remaining 1900K instances. We believe that in the future, unification of often ambiguous units (e.g. cups, pinch) might have a wide range of applications in and outside of the culinary world, as well as additional unification utilizing knowledge graphs, is another exciting future project. As the dataset is large and organized by genre, medical sectors, particularly those working with food nutrition, can recommend a variety of meals to the patients. If the recipe's portion can be estimated, a large area will open up, which is the components calories, which can be used to analyze food calories intake for various types of food analysis or nutrients. Finally, by training these massive datasets, an application can be created to build a new menu and generate buzz in the food market, giving consumers a new taste and direction to manufacture such delicacies, which may be a big contribution to the culinary sector.
%
% ---- Bibliography ----
%
% BibTeX users should specify bibliography style 'splncs04'.
% References will then be sorted and formatted in the correct style.
%
\bibliographystyle{splncs04}
\bibliography{bib}
\end{document}